\documentclass[conference]{IEEEtran}  

\IEEEoverridecommandlockouts                              
\usepackage{caption}

\usepackage{amssymb}
\usepackage{amsmath}
\usepackage{algorithm, algorithmic}
\usepackage{multirow}
\usepackage[tight,scriptsize]{subfigure}
\usepackage{setspace}
\usepackage{graphicx}

\begin{document}

\title{Resource-Efficient Computing in Wearable Systems\\
}

\author{
    \IEEEauthorblockN{Mahdi Pedram\IEEEauthorrefmark{1}, Mahsan Rofouei\IEEEauthorrefmark{2}, Francesco Fraternali\IEEEauthorrefmark{3}, Zhila Esna Ashari\IEEEauthorrefmark{1}, Hassan Ghasemzadeh\IEEEauthorrefmark{1}}
    \IEEEauthorblockA{\IEEEauthorrefmark{1}Electrical Engineering and Computer Science, Washington State University
    \\\{mahdi.pedram, z.esnaashariesfahan, hassan.ghasemzadeh\}@wsu.edu}
    \IEEEauthorblockA{\IEEEauthorrefmark{2}Google 
    \\\ rofouei@gmail.com}
    \IEEEauthorblockA{\IEEEauthorrefmark{3}Computer Science and Engineering, University of California San Diego
    \\\ frfrater@eng.ucsd.edu}
}

\maketitle
\thispagestyle{empty}
\pagestyle{empty}

\begin{abstract}
We propose two optimization techniques to minimize memory usage and computation while meeting system timing constraints for real-time classification in wearable systems. Our method derives a hierarchical classifier structure for Support Vector Machine (SVM) in order to reduce the amount of computations, based on the probability distribution of output classes occurrences. Also, we propose a memory optimization technique based on SVM parameters, which results in storing fewer support vectors and as a result requiring less memory. To demonstrate the efficiency of our proposed techniques, we performed an activity recognition experiment and were able to save up to $35\%$ and $56\%$ in memory storage when classifying $14$ and $6$ different activities, respectively. In addition, we demonstrated that there is a trade-off between accuracy of classification and memory savings, which can be controlled based on application requirements.


\end{abstract}

\IEEEpeerreviewmaketitle

\section{Introduction}
\vspace{-1mm}
Emerging embedded wireless sensor systems are targeting a broad range of applications such as on-body monitoring systems. Examples of such systems are activity logging systems \cite{11}, sleep monitoring devices \cite{milici2018wireless} and on-body temperature measurement systems \cite{15}. Although the signals measured from these sensor systems contain valuable information, they require certain amount of processing, memory and power for interpreting these signals and detecting a specific condition. Many of the emerging applications that benefit from such systems require real-time interpretation of sensor measurements. 

Due to the limitations such as storage and processing power, most of current embedded sensor systems assume that data is transferred to a base-node for offline processing. However, some emerging applications require resource-efficient algorithms that can run real-time. As body-worn sensor systems are becoming more pervasive, local processing is becoming desirable because of avoiding interference effects from other radio signals transmitting data. Therefore, it is beneficial in terms of reliability. Also, time constraints imposed by application needs, is crucial in the execution of tasks. A small time delay may cause malfunctioning or even failure in execution. In addition, communication system is the main culprit to consume most of the energy in wearable sensors \cite{mukhopadhyay2015wearable}. Therefore, local processing is also beneficial in terms of minimizing the amount of traffic.

Low energy consumption is one of the key design goals of the current embedded sensor systems. Typically programmable processors are the core of such systems. Power analysis of these processors indicates that a significant amount of power is consumed in the on-chip (instruction) memory hierarchy ~\cite{16}. Also, minimizing memory requirements has direct impact on systems performance, power dissipation, reducing the size and overall cost of an embedded system ~\cite{1}. Thus, from another perspective reducing the number of instructions executed based on real-time events lowers the system overall power consumption.

In addition to application-specific requirements, real-time annotated classified data can also enable further power saving mechanisms that include turning off sensor nodes which are not needed based on activity being performed. Also, classified data enables memory saving in this way that only annotated features can be stored instead of complete raw signals. Furthermore, real time classification can reduce the amount of processing based on specific activities required processing. 
For instance, a fall detection application or a gait cycle detection such as \cite{wu2015development} and \cite{ma2019cyclepro} are interested in certain gait features to predict events early enough for alerting the user. Therefore, user activities need to be classified in real-time and before a deadline to enable this. Thus, real-time classification can be used for detecting context (e.g. Physical activity), in ubiquitous context-aware applications. 

In this work, we use Support Vector Machine (SVM)~\cite{105}, as a supervised learning framework for interpreting real-time measured signals and classifying states. SVMs are in wide-spread use and are popular in medical applications mainly because of their robustness when minimal training data is available. Based on the above discussion, our contribution in this work is twofold: First, we derive a hierarchical classifier model using Support Vector Machines (SVM) that inherently reduces the amount of computation by classifying events more likely to happen, earlier in the decision path while at the same time guaranteeing to meet time constraints for classification. Our second contribution is a memory optimization technique that organizes classifier parameters and results in requiring less memory for classifier implementation. 

\section{Related Work}\label{section_related}
\vspace{-1mm}
Advances in bio-engineering have led to increasing number of systems that require real-time classification of bio-signals. For example, real-time classification of EMG signals for prosthetic devices for paralyzed individuals~\cite{9}. Another example is real-time classification of ECG data for detecting heart rhythm irregularities~\cite{datta2017identifying}. Other systems are ones that classify different states of the body by monitoring various physiological measurements with applications to fall detection, energy expenditure calculations and etc. ~\cite{11}. 


One of the limitations of on-line classification on nodes is memory requirements. In order to reduce energy consumption of the memory subsystem in embedded systems, researchers have investigated  the ways to decrease the energy needed for both instruction and data memory. For decreasing instruction memory energy, several approaches have been suggested including reducing memory access count~\cite{12} and reducing bus activity~\cite{13}. The work in~\cite{12} applies instruction compression to reduce memory access count. Several approaches have also been suggested for reducing data memory energy, including loop transformation and data layout optimization~\cite{1}.  

In this work, we use Support Vector Machine (SVM)~\cite{105} to perform activity recognition. The SVM classifier approach is a popular choice specifically in activity recognition ~\cite{201,202}. The work in~\cite{201} presents an SVM-based classification approach that achieves 98\% average accuracy for classifying six different postures and activities. A multimodal physical activity recognition system is developed in~\cite{202} by fusing both ambulatory Electrocardiogram (ECG) and accelerometer information together to reach a classification accuracy of 97.3\%.

There has been several attempts to enable real-time classification using SVMs on embedded systems~\cite{103}, ~\cite{chen2017robust}. ~\cite{103} focuses on proposing new a approach for implementing SVM on digital architectures such as FPGAs. Lee, et. al.~\cite{200} present a formulation for kernel function of a SVM classifier. Their proposed changes result in reductions in the amount of real-time computations required for classification. However the changes in~\cite{200} only apply to kernel functions employing polynomial transformations. 

As mentioned in ~\cite{200}, the SVM model can be derived offlineand thus, the energy for its training is not of primary concern. Therefore, the main concern, which is the focus of this manuscript, is optimizing the real-time classification process by preparing a set of memory-optimized support vectors off-line and use them in real-time classification.

\vspace{-1mm}

\section{Problem Statement}\label{section_problem}
In this section we propose the problem of constructing our classification model based on SVM. The classifier's type and configuration are determined using the constraints of the real-time systems, such as time and power limitations. In addition, we formulate the problem of storage minimization on the basis of the designed classification architecture and prove the complexity of the problem.   

\subsection{Classification Model}
In this manuscript, we have selected a variation of SVM, that is Hierarchical classifier. SVM is primarily designed for binary classification problems. However, in order to classify multi-class problems, new structures are required. Generally, multi-class classification problems are decomposed into many binary class problems arranged in a structure called Hierarchical Classifiers.

In the hierarchical classification approach, multiple classifiers are constructed at different levels. First off, an activity is classified at the top level and it is classified to one or more lower levels. This process continues until all activities have been classified. The structure of the hierarchical classifier can be constructed in various ways and one of the most basic ones is using tree structure. A specific form of a tree structure classifier is the Binary Hierarchical Classifier (BHC) described in~\cite{3}. Another known structure is a Directed Acyclic Graph (DAG) structure in which a node can have more than one parent, as opposed to tree structure. Based on these two generic structures, more specific structures have been built and discussed, such as one-against-all~\cite{7}, and one-against-one~\cite{8} classifiers. In both of these structures leaf nodes represent activities, while internal nodes represent classifiers. Therefore, based on this definition, an n-class classifier has $n$ leaves. ~\cite{4} describes an example of the DAG structure SVM classifier.

There are different ways of constructing a BHC or a DAG classifier for a n-class classification problem. For example the number of different binary trees on n nodes is Cn, the nth catalan number, which equals to $\frac{1}{n+1}\binom{2n}{n}$. Also, there are $n(n-2)!$ different ways of constructing the DAG. However, selecting one of these structures depends on many factors. 

In our model, we choose the structure of the classifier using the statistical information on probability of activities occurrences within an application. In this way, by choosing the activity with highest probability to be placed on the top (root) of the tree, the expected amount of computations required for classification of different activities is minimized. 

\subsection{Problem Formulation}
\subsubsection{Main Problem}
In this manuscript, we propose a solution to the following problem.
\newline
\textbf{Problem 1.} \textit{Given a set of $n$ possible activities $A= \{ A_{1}, A_{2}, .., A_{n} \}$ with a probability distribution of $P=\{ P_{1}, P_{2}, ..., P_{n} \}$ and a time constraint $T= \{T_{1}, T_{2}, .., T_{n}\}$ for classifying each activity, the task is to classify real-time sensor data to one of the activities in $A$, meeting time constraint $T$, while optimizing  memory usage and the amount of processing.}
\newline

\vspace{-3mm}

Time constraints imposed by application needs, is often crucial in the execution of tasks and a small time delay may cause malfunctioning or missing real-time events.
Thus in this problem, $T$ represents the deadlines for classification of each activity. 

We break the stated problem into two different sub-problems: finding a classifier that meets the $T$ time constraint for all paths of the hierarchical classifier; And performing memory optimization on the derived classification structure from the first part, to achieve a memory-efficient design, directly impacting system's performance and power dissipation.

The algorithm designed for acquiring a unique BHC configuration for the first sub-problem, is proposed in Section~\ref{section_struct}. This configuration will be the input to the second sub-problem. Then, we propose our designed algorithm for memory optimization in Section~\ref{section_memmin}.

In the rest of this section, we elaborate on the formulation of the second sub-problem. 

\subsubsection{Memory Optimization Problem}
Consider a simple SVM classifier, used for classification of events of interest. Based on Equation~\ref{eq_1} \cite{105} a set of support vectors extracted through the training phase need to be stored in the memory of the sensor node. 

\vspace{-2mm}
\begin{equation}
f(x)=\sum \alpha_{i}.y_{i}.x_{i}^{T}.x + b
\label{eq_1}
\end{equation}

These support vectors are then used to perform classification in real-time on the sensor node. Therefore, the key to minimize memory requirements for implementation of a hierarchical classifier is reducing the number of support vectors needed to be stored at each level. Solely reducing the number of support vectors, will result in decrease of classification accuracy at each level and then in the overall hierarchical classifier. However, substitution of support vectors with vectors within certain distances from them, might be beneficial while maintaining a lower-bound on the accuracy of the hierarchical classifier. 

The whole idea is that by substitution of vectors, we can derive a set of support vectors for each layer in the hierarchical classifier in such a way that some support vectors are shared within different layers. In other words, we suggest replacing support vectors with vectors near them in such a way the number of overlapping support vectors between layers of the hierarchical classifier are maximized. With this approach, we can minimize the amount of storage required for storing support vectors while at the same time maintaining a certain accuracy level. 

Our memory optimization technique runs on a set of $n$ Input Support Vectors ($ISV_{i}$) extracted from an initial run of a hierarchical classifier and results in $n$ Final Support Vectors ($FSV_{i}$).
\newline
\textbf{Definition 1.}\textit{(SECONDARY SUPPORT VECTOR): Vector $v_{i}$ is considered as a Secondary Support Vector ($SSV$) of $SV_{i}$ if:}
\begin{equation}
d(v_{i},SV_{i})\leq \varepsilon 
\label{eq_2}
\end{equation}
\textit{Where $d(a,b)$ represents the Euclidean distance of a and b.}
\newline

\vspace{-2mm}

\textbf{Definition 2.}\textit{(OVERLAPPING SUPPORT VECTOR): Vector $v_{i}$ is considered as an Overlapping Support Vector (OSV) if $v_{i}$ is a Secondary Support Vector of $SV_{l}$ and $SV_{t}$ where $l\neq t$ and $v_{i}$ is chosen as a final support vector.}

At the intuitive level, Maximal Overlap Classification problem may be defined as the selection of final support vectors in such a way that total number of overlapping vectors is maximized. We can formulate a simplified version of the Maximal Overlapping Classification (MOC) problem in the following way:

\textbf{Problem 2.} \textit{MAXIMAL OVERLAPPING CLASSIFICATION (MOC): Given a finite set $ISV= {ISV_{1}, ISV_{2}, ...ISV_{n}}$ of initial support vectors from running an initial SVM hierarchical classifier of $c$ classifiers, a finite set of $SSV= {SSV_{1}, SSV_{2},... SSV_{m}}$, consisting of all Secondary Support Vectors, the task is to select a set of $n$ FSVs in such a way that the number of Overlapping Support Vectors are maximized while achieving a minimum bound of $CA$ percent on the classifier accuracy.}
\newline
In order to clarify Problem 2, assume $a_{ij}$ is a given binary that determines if vector $i$ is a secondary support vector of $ISV_{j}$.

\vspace{-3mm}

\begin{equation}
a_{ij}=\begin{cases}
1 & \text{ if vector i is a SSV for}  ISV_{j}  \\ 
0 & \text{ } o.w.
\end{cases}
\label{eq_3}
\end{equation}

and $x_{i}$ is a binary variable that indicates whether or not secondary support vector $i$ is selected as a final support vector.
\vspace{-4mm}

\begin{equation}
x_{i}=\begin{cases}
1 & \text{ if }  SSV_{i} \text{ is selected as a FSV }  \\ 
0 & \text{ } o.w.
\end{cases} 
\label{eq_4}
\end{equation}

\vspace{-1mm}
Below is the corresponding Integer Linear Programming (ILP) formulation of the MOC problem (Problem 2). 

Objective:	
\begin{equation}
Minimize \sum_{i=1}^{m+n}x_{i}
\label{eq_5}
\end{equation}

\vspace{-2mm}

Subject to:   
\vspace{-2mm}
\begin{equation}
Minimize \sum_{i=1}^{m+n}a_{ij}x_{i}\geq 1\; \; \; \; \; \text{for all j: }1\leq j\leq n
\label{eq_6}
\end{equation}

\vspace{-3mm}

\begin{equation}
x_{i}\in \left \{ 0,1 \right \}\; \; \; \; \; \; \; 1\leq i \leq m+n 
\label{eq_7}
\end{equation}

\subsection{Problem Complexity}
\vspace{-1mm}
In this section, we provide problem complexity analysis for the Maximal Overlapping Classification (MOC) problem. In order to prove that the MOC problem is NP-complete, we transform an instance of Hitting Set problem, into MOC. Hitting Set problem is the dual of Set Cover and is NP-complete. An instance of the classic Set Cover problem can be viewed as a bipartite graph, where sets and elements of the universe represent left and right vertices respectively and edges show the inclusion of elements in sets. In the Hitting Set problem, the objective is to cover the left vertices (sets) using a minimum subset of the right vertices (elements).
By considering each classifier in the hierarchical classifier tree as a set and each vector $v_{i}$ in $V$ as an element, the MOC problem becomes the problem of covering each classifier using a minimum subset of vectors, which is solved by Set Cover or Hitting Set. Therefore, it can be concluded that MOC problem is NP-complete.

\vspace{-2mm}
\section{Memory-Efficient BHC}\label{section_mem}
\vspace{-1mm}
In this section, we provide our proposed solution for Problem 1 in two parts. In the first part in Section~\ref{section_struct}, we describe how we perform classifier structure extraction for a set of activities with the given probability distribution meeting time constraint $T$. The result of this section is a Binary Hierarchical Classifier with a unique structure. Then, in Section~\ref{section_memmin}, we perform memory optimization based on the classification structure extracted in Section ~\ref{section_struct}. 

\subsection{Classifier Structure Extraction}\label{section_struct}
\vspace{-1mm}
There are many different possible structures for a binary tree of $n$ nodes. However, the choice of tree structure is crucial in meeting the constraints of the system. Instead of using one of the generic classifier structures such as one-against-all, we show how the BHC tree structure can be chosen in order to minimize the expected number of instructions executed in run-time, while at the same time meeting system timing constraints.

\vspace{-4mm}

\begin{equation}
E(I)=\sum_{i=1}^{n}p_{i}l_{i}
\label{eq_8}
\end{equation}

\vspace{-1mm}
 	 	
Algorithm~\ref{alg1} which is based on Huffman coding~\cite{5} finds the optimal tree structure for the BHC such that the expected number of instructions ($I$) for a given activity determined in run-time is minimized (Equation~\ref{eq_8}). 


\begin{algorithm}[h]
\begin{algorithmic}[1]
\STATE {\bf Input:} Activities $A= \{A_{1}, A_{2}, ..., A_{n} \}$
\STATE {\bf Input:} Corresponding probability distribution of $P=\{ P_{1}, P_{2}, ..., P_{n}\}$
\STATE {\bf Input:} Time constraint of $T= \{T_{1}, T_{2}, ..., T{n} \}$
\STATE {\bf Output:} A unique Tree structure
\STATE Construct a forest $N = \{N_{1}, N_{2}, ..., N_{n}\}$ of n binary trees of only one node with empty left and right children. The value of each node is its corresponding Pi.
\STATE Select two trees Ni and Nj with minimum probability values from P to construct a new binary tree $N_{y}$ with $N_{i}$ and $N_{j}$ as its children. The probability of the root is the sum of $P_{i}$ and $P_{j}$.
\STATE $N \leftarrow N  - \{Ni , Nj \}$
\STATE $N \leftarrow N  U \{ Ny \}$
\STATE Repeat steps 6-8 until $|N|=1$ which contains the resulting tree.
\end{algorithmic}
\caption{Classification Tree Decomposition}
\label{alg1}
\end{algorithm}


Here $l_{i}$ represents distance from root ($l_{i}=0$ for root). We assume that the number of executed instructions is proportional to the number of classifiers. Therefore, the number of instructions executed for an activity at $l_{i}=2$ is proportional to $2$ since it has to pass through two classifiers (one at the root and one at $l_{i}=1$). 

Algorithm~\ref{alg1} finds an optimal tree structure for BHC, but does not necessarily satisfy timing constraints. Based on the above stated assumption that the number of instructions is proportional to the number of classifiers, time of execution is also proportional to the number of classifiers and in turn to the depth of the tree. Therefore, to meet system timing constraints, the depth of the BHC should not exceed $a$ $T_{min}$, where $T_{min}$ is the earliest time constraint in $T$ and $a$ is a constant parameter. 

One of the known algorithms for producing Huffman codes with a constraint on code length is the Package-Merge algorithm described in~\cite{6}, which is an $O(nL)$ algorithm where $L$ is maximum code length. We modify the package-merge algorithm in order to meet system timing constraints for classification. This algorithm consists of two parts of package and merge. In the package step an item at level $i$ is constructed by merging two items at level $i-1$. We use Algorithm~\ref{alg1} for the packaging step and keep the merge step unchanged. Details of the package-merge algorithm are in~\cite{6}. The time complexity of the classification tree decomposition considering constraints on the depth of the classifier is $O(nlogn.L)$.

\vspace{-1mm}

\subsection{Memory Minimization}\label{section_memmin}
\vspace{-1mm}

Algorithm~\ref{alg2} represents our proposed algorithm for memory minimization. 

\begin{algorithm}[h]
\begin{algorithmic}[1]
\STATE {\bf Input:}  A unique BHC from Algorithm~\ref{alg1}, A set of n initial support Vectors ($ISV$s), $\epsilon$: maximum distance from ISV.
\STATE {\bf Output:} A set of n Final Support Vectors ($FSV$s).
\STATE Construct the set $SSV= {SSV_{1}, SSV_{2}, ... SSV_{m}}$,  of all secondary support vectors.
\STATE $V \leftarrow$ Set of all vectors $v_{i}$ where $vi \in ISV$ or $vi \in SSV$. (n+m elements).
\STATE Sort set V in order of frequency of overlapping initial Support vectors (ISVs).
\STATE $FSV \leftarrow {}.$
\STATE Select a vector $vi \in V$ that maximizes $ | vi  \cap ISV|$
\STATE $ISV \leftarrow ISV$ - {corresponding ISV elements covered by $v_{i}$ }.
\STATE $FSV \leftarrow FSV \cup { vi }.$
\STATE Repeat steps 7-9 until ISV is empty.
\end{algorithmic}
\caption{Greedy Memory Optimization}
\label{alg2}
\end{algorithm}

\vspace{-4mm}


The outcome of Algorithm~\ref{alg2} is a set of $n$ Final Support Vectors ($FSV$). This set is initially empty (line 6) and is constructed by selecting $v_{i}$ vectors from Set $V$ which is initially composed of a sorted list of all Initial Support Vectors and Secondary Support Vectors (lines 4-5). At every step, a vector $v_{i}$ is selected from $V$ in such a way that as many possible uncovered $ISV$s are covered (the number of overlaps are maximized). In other words the vector which covers/overlaps with most of $ISV$s is chosen as a $FSV$. The corresponding $ISV$s which are covered by $v_{i}$ are eliminated from $ISV$ (lines (7-9). The selection procedure continues until $n$ Final Support Vectors are chosen (line 10).

\vspace{-1mm}

\section{Experimental Results}\label{section_exp}
\vspace{-1mm}

\subsection{Experimental Set-up}\label{section_setup}
In order to demonstrate the efficiency of our proposed methods, we performed an experiment. A set of $14$ different activities were performed by $3$ healthy subjects with the average age of $27$ years old. Sensor measurements were collected from $7$ sensor nodes placed on different body locations, comprising of a 3-axis accelerometer and 2-axis gyroscope with sampling rate of $50$Hz. Sensor locations are waist, right wrist, left wrist, right arm, left thigh, right ankle and left ankle; selected based on having diversity in samples, keeping most informative ones and ignoring the highly correlated samples gained from some symmetric body parts. The $14$ activities performed by subjects are: stand to sit(A1), sit to stand(A2), sit to lie(A3), lie to sit(A4), jump(A5), turn clockwise(A6), grasp object from ground(A7), rise to bend(A8), step forward(A9), step backward(A10), look back(A11), Kneel(A12), rise from kneeling(A13), and return from looking back(A14). Then, we developed a Binary Hierarchical Classifier, optimized  for both memory and expected value of instructions, while satisfying classification time constraints. 

\vspace{-1mm}
\subsection{Results}\label{section_res}

\begin{figure}[h]
\centering
\includegraphics[width=0.99\linewidth]{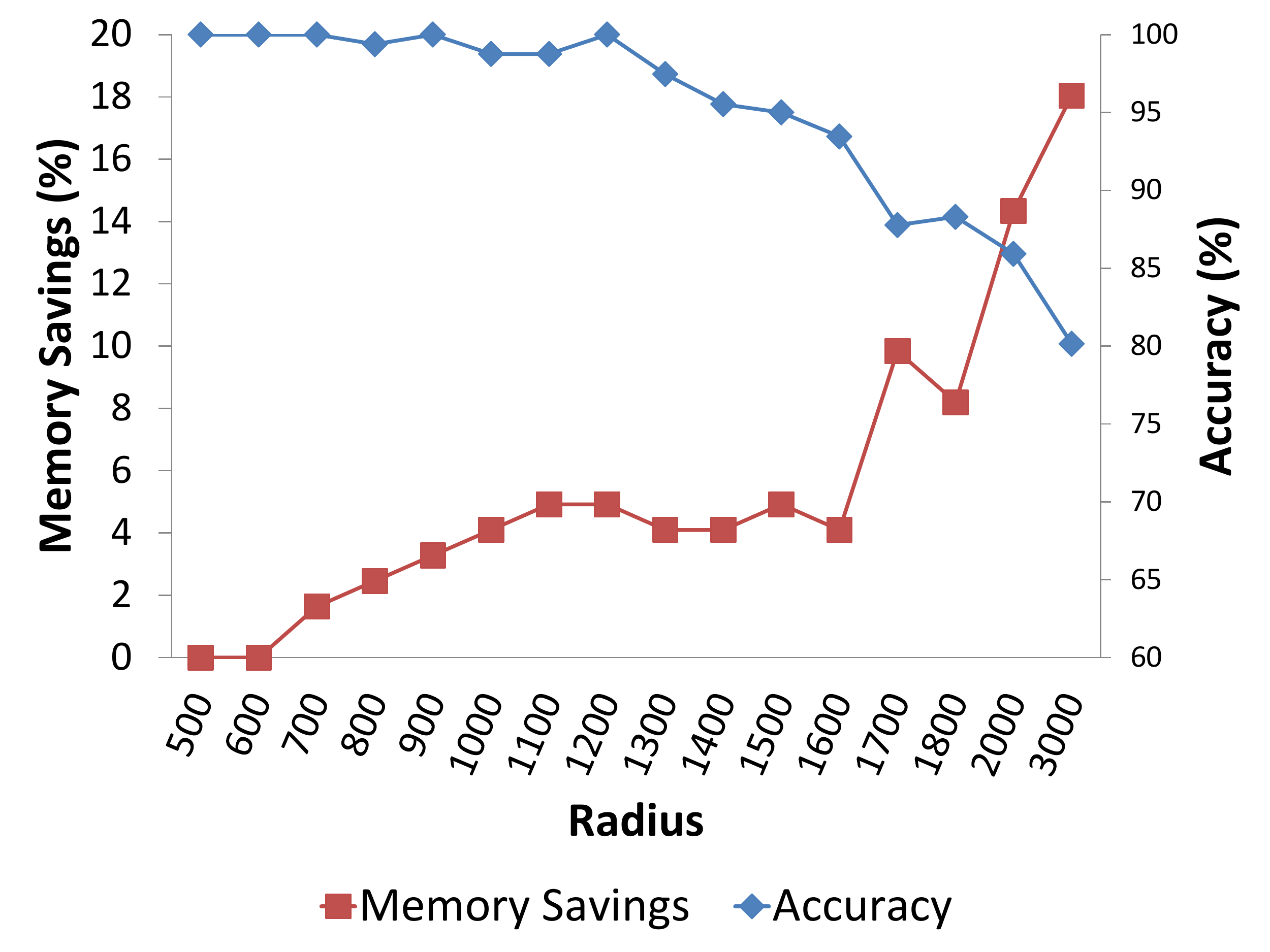}
\caption{Memory Savings and Classification Accuracy vs. Radius}

\label{fig5}
\end{figure}

We first focus on efficiency of our memory optimization technique. Figure~\ref{fig5} shows the results for using Algorithm~\ref{alg2} on the dataset of $14$ activities, while the memory optimization is performed on a derived tree structure from Algorithm~\ref{alg1}. This figure shows the amount of memory savings for different values of $\varepsilon$, determining the number of SSVs, with respect to the memory usage of the initial classifier before optimization. It also includes their corresponding classification accuravy values. It can be seen that in the best case, we achieve up to 35\% memory savings with a classification accuracy of 60\%.  


As Figure~\ref{fig5} suggests, when we have increase in memory savings, the accuracy of classification decreases. This is due to the fact that with increase in $\varepsilon$, the number of secondary support vectors for each initial support vector increases and this in turn increases the chances of finding more overlapping support vectors between different ISVs and results in a decrease in classification accuracy. Therefore, there is a trade-off between memory savings and accuracy and based on application constraints and requirements, different choices of $\varepsilon$ can be made. For instance, to avoid significant decrease in accuracy for medical applications, we need to limit memory savings.

In the next step we show the effect of tree structure selection, as the output of Algorithm~\ref{alg1}, in overall savings both in terms of memory optimization and reducing the number of executed instructions. We have explored various tree structures based on probability distributions for a smaller subset of activities consisting of $6$ activities of $A_{1}$, $A_{3}$, $A_{5}$, $A_{6}$, $A_{7}$, $A_{9}$. We constructed this subset in order to test this part on a different dataset compared to last part; and also to control the the depth of hierarchical tree structures. However, to include more diversity in the experiment, five different randomly generated probability distributions have been assigned to the dataset as input $P$ of algorithm. Table~\ref{tbl2} shows various information for the classifier structure generated based on given probability distributions such as depth of the tree (which needs to satisfy system timing constraints), $E(I)$ (Expected number of Instructions executed at run-time and calculated by equation $8$), and initial memory overlap (indicating the amount of Overlapping Support Vectors found in the initial Hierarchical classifier without performing memory optimization).


\begin{table}[h]
\begin{center}
\centering
\caption{Tree Decomposition Results} 
\label{tbl2}
\begin{tabular}{|p{4.25cm}|p{0.75cm}|c|p{1cm}|}
\hline
\centering
\textbf{Probability \\
Distribution $\{A_{1} ,A_{3} ,A_{5} ,A_{6} ,A_{7} ,A_{9}\}$} &
\textbf{Depth of Tree} & \textbf{$E(I)$} & \textbf{Overlap
(\%)} \\ \hline
$P_{1}=\{20 ,20 ,5 ,5 ,10 ,40\}$ &	5 &	230 &	33.33 \\ \hline
$P_{2}=\{33 ,5 ,10 ,10 ,12 ,30\}$ &	3 &	282 &	42.42 \\ \hline
$P_{3}=\{10 ,13 ,42 ,10 ,5, 20\}$  &	4 &	234 &	28 \\ \hline
$P_{4}=\{10 ,10,15,15,25,25\}$ &3 &	250 &	30 \\ \hline
$P_{5}=\{8 ,10 ,7 ,12 ,21 ,42\}$ &	4 &	231 &	34.48 \\ \hline
\end{tabular}
\end{center}
\end{table}

\vspace{-3mm}

\begin{figure}[h]
\centering
\includegraphics[width=0.99\linewidth]{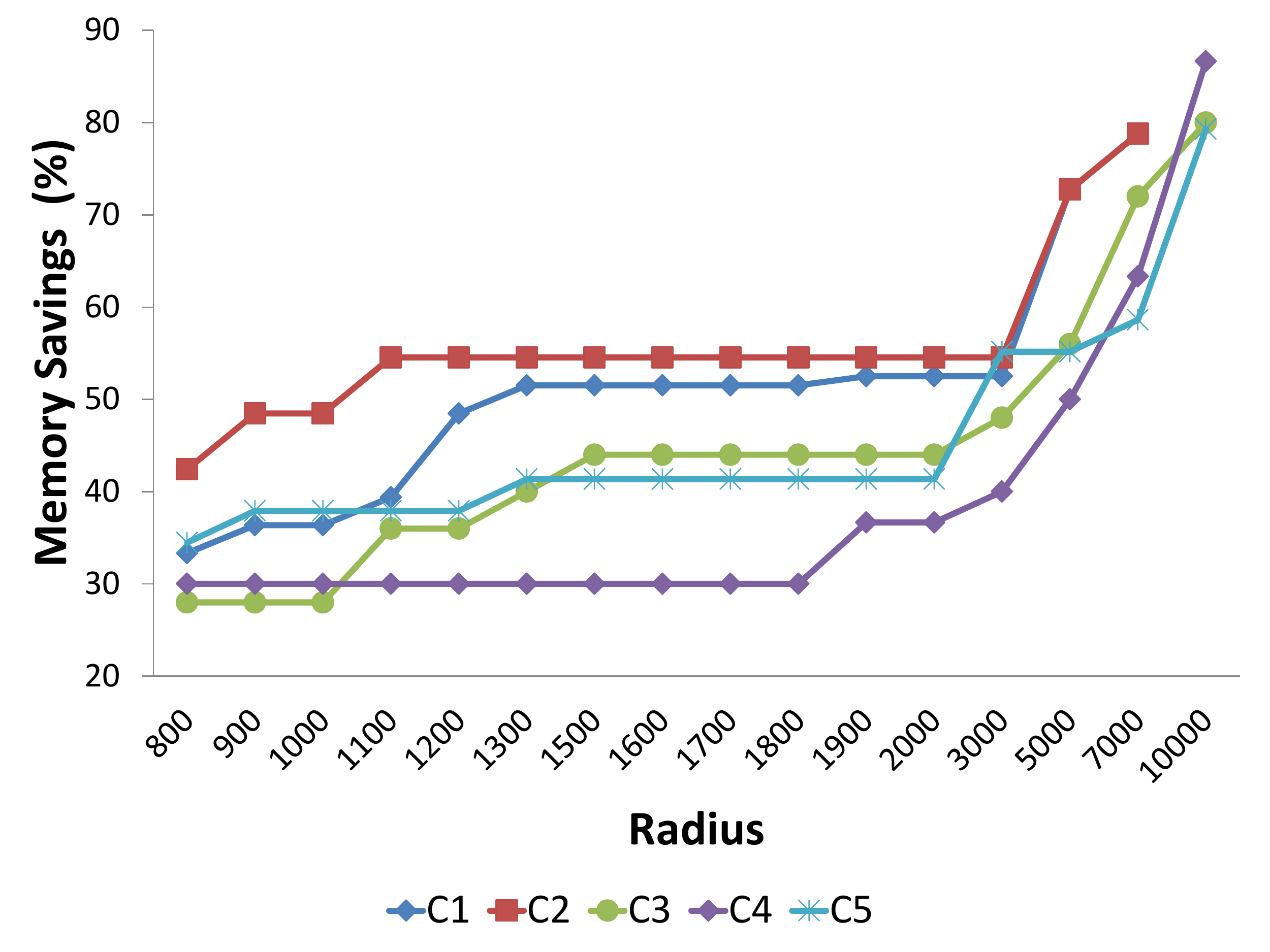}
\caption{Memory Savings vs. Radius for Five Classifiers. }
\label{fig6}
\end{figure}

\begin{figure}[h]
\centering
\includegraphics[width=0.99\linewidth]{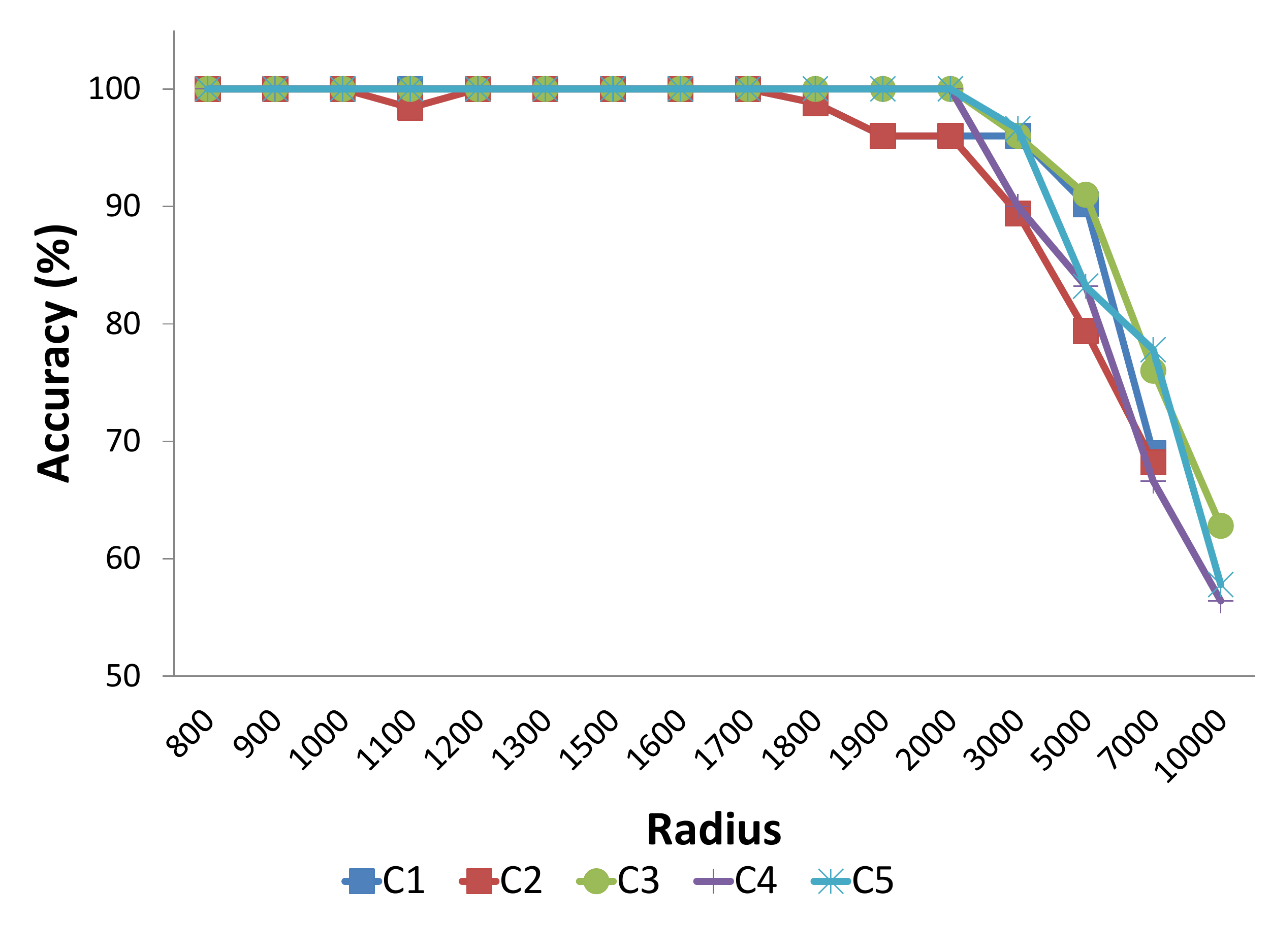}
\caption{Classification Accuracy vs. Radius for Five Classifiers.}
\label{fig7}
\vspace{-4mm}
\end{figure}

Figure~\ref{fig6} presents memory optimization results with respect to initial savings for the $5$ different classifiers in Table~\ref{tbl2}, showing the effect of initial tree structure selection on memory savings. As mentioned, the results are for the same $6$ activities but with different probability distributions. Also, Figure~\ref{fig7} presents their corresponding classification accuracy numbers. 

\vspace{-2mm}

From figures~\ref{fig6} and~\ref{fig7}, it is seen that up to $56.6$\% memory savings can be achieved for classifier $4$, when accuracy is $56$\%. Also, while maintaining a $100$\% accuracy, as the upper-bound, we can achieve $18.1$\% memory savings for classifier $1$. Note that the subset of $6$ activities includes more distinct activities compared to the original set of $14$ activities which results in higher accuracy for upper-bound. The achieved results confirm the capability of our proposed methods in saving memory in real-time classification systems. In addition, from these two figures, the trade-off between accuracy and memory efficiency is concluded, which is in accordance with figure~\ref{fig5}, and can be controlled based on applications requirements. 

\section{Conclusions}\label{section_conc}
\vspace{-1mm}
We presented two optimization techniques for real-time signal classification in lightweight embedded systems. We proposed a technique to extract the hierarchical classifier structure based on statistical information about occurrences of various events. This method minimizes the expected amount of required computations while meeting system timing constraints. In addition, we proposed a memory optimization technique that results in a memory-efficient implementation of the multi-class SVM classifier for embedded systems. We demonstrated the efficiency of our proposed algorithms using the datasets collected from $3$ subjects performing $14$ and $6$ different activities, and we could achieve up to $35\%$ and $56\%$ of saving the memory respectively. Also, we demonstrated that there is a trade-off between accuracy of classification and memory savings, which can be controlled based on application requirements.  

\section*{Acknowledgment}
\vspace{-1mm}
This work was supported in part by the United States Department of Education, under Graduate Assistance in Areas of National Need (GAANN) Grant P200A150115, and the United States National Science Foundation, under grant CNS-1750679. Any opinions, findings, conclusions, or recommendations expressed in this material are those of the authors and do not necessarily reflect the views of the funding organizations.



\bibliography {mybib}
\vspace{-2mm}
\bibliographystyle {IEEEtran}



\end{document}